\documentclass{article}
\usepackage[preprint]{neurips_2019}

\usepackage[utf8]{inputenc} 
\usepackage[T1]{fontenc}    
\usepackage{hyperref}       
\usepackage{url}            
\usepackage{booktabs}       
\usepackage{amsfonts}       
\usepackage{nicefrac}       
\usepackage{microtype}      
\usepackage{times}
\usepackage{soul}
\usepackage[small]{caption}
\usepackage{graphicx}
\usepackage{amsmath}
\usepackage{algorithm}
\usepackage{algorithmic}
\usepackage{xspace}
\usepackage{xcolor}
\usepackage{listings}
\usepackage{epsfig}
\usepackage[title]{appendix}
\usepackage{makecell}
\usepackage{multirow}
\usepackage{enumitem}
\usepackage{subcaption}

\DeclareFixedFont{\ttb}{T1}{txtt}{bx}{n}{8}
\DeclareFixedFont{\ttm}{T1}{txtt}{m}{n}{8}

\usepackage{color}
\definecolor{deepblue}{rgb}{0,0,0.5}
\definecolor{deepred}{rgb}{0.6,0,0}
\definecolor{deepgreen}{rgb}{0,0.5,0}
\definecolor{purple}{rgb}{0.5,0,0.5}
\definecolor{gray}{rgb}{0.33,0.33,0.33}

\definecolor{dkgreen}{rgb}{0,0.6,0}
\definecolor{gray}{rgb}{0.5,0.5,0.5}
\definecolor{mauve}{rgb}{0.58,0,0.82}

\newcommand{\preeq}{\vspace{0mm}\begin{small}}
\newcommand{\posteq}{\vspace{0mm}\end{small}}

\title{Making Classical Machine Learning Pipelines Differentiable: A Neural Translation Approach}

\author{
	Gyeong-In Yu{\normalfont \textsuperscript{1}}, Saeed Amizadeh{\normalfont \textsuperscript{2}}, Sehoon Kim{\normalfont \textsuperscript{1}}, Artidoro Pagnoni{\normalfont \textsuperscript{3}},\\
	{\bf Byung-Gon Chun}{\normalfont \textsuperscript{1}}, {\bf Markus Weimer}{\normalfont \textsuperscript{2}}, {\bf Matteo Interlandi}{\normalfont \textsuperscript{2}}\\
	\textsuperscript{1}Seoul National University, \textsuperscript{2}Microsoft,
	\textsuperscript{3}CMU\\
	\texttt{\{gyeongin,sehoon95,bgchun\}@snu.ac.kr},\\
	\texttt{\{saamizad,mweimer,mainterl\}@microsoft.com},\\
	\texttt{apagnoni@andrew.cmu.com}
}

\begin{document}

\makeatletter
\newcommand\notsotiny{\@setfontsize\notsotiny{9.0}{11}}
\newcommand\tablecolumnmarginsmall{\setlength\tabcolsep{3.5pt}}
\makeatother

\maketitle

\begin{abstract}
Classical Machine Learning (ML) pipelines often comprise of multiple ML models where models, within a pipeline, are trained in isolation.
Conversely, when training neural network models, layers composing the neural models are simultaneously trained using backpropagation.
We argue that the isolated training scheme of ML pipelines is sub-optimal, since it cannot jointly optimize multiple components.
To this end, we propose a framework that translates a pre-trained ML pipeline into a neural network and fine-tunes the ML models within the pipeline jointly using backpropagation.
Our experiments show that fine-tuning of the translated pipelines is a promising technique able to increase the final accuracy.
\end{abstract}

\section{Introduction}
\label{sec:intro}

Deep Neural Networks (DNNs) have been exceptionally successful in pushing the limits of various fields such as computer vision and natural language processing~\cite{imagenet,parity}.
Nevertheless, classical Machine Learning (ML) techniques such as gradient boosting and linear models are still popular among practitioners~\cite{kaggle-survey}, especially because of their intrinsic efficacy and interpretability.
When using these techniques, we often build a \textit{machine learning pipeline} by composing multiple data transforms and ML models~\cite{mldotnet,scikit}.
This abstraction allows users to capture the data transformation pipeline as Directed Acyclic Graphs (DAGs) of operators.

Many of the top performing ML pipelines in the industry and Kaggle's competitions~\cite{kaggle-criteo,komaki}
often include more than one \textit{trainable operator,} i.e., ML models or data transforms that determine how to process input by learning from the training dataset.
These trainable operators are often trained sequentially by following the topological order specified in the DAG.
In this paper, we claim that this sequential training of ML pipelines' operators is sub-optimal since the operators are trained in isolation and are not jointly optimized.
This approach substantially differs from how DNNs are trained.
DNN layers, which can also be seen as multiple cascaded operators, are typically trained \textit{simultaneously} using \textit{backpropagation} by which parameters can be globally estimated end-to-end to reach better (local) minima.
Arguably, this is one of the most fundamental features of deep learning.

Inspired by these observations, we propose an approach whereby (possibly) trained ML pipelines are translated into neural networks and fine-tuned therein.
By doing so, we can use backpropagation over ML pipelines in order to bypass the greedy one-operator-at-a-time training model and eventually boosting the accuracy of the entire ML pipeline.
During the translation phase, we can retain the information already acquired by training the original ML pipeline and provide a useful parameter initialization for the translated neural network, making the further training of the network more accurate and faster.

Nevertheless, noticeable challenges arise when translating pipelines involving data transforms or models, such as word tokenization or decision tree, that are intrinsically non-differentiable.
We propose neural translation for selected non-differentiable operators including decision tree, one-hot encoding, and Latent Dirichlet Allocation.
We also suggest controlling which part of the neural network (translated from a decision tree) to be further trained as a natural way of setting the trade-off between fit and bias.

We conduct experiments on three real world datasets over various pipeline configurations containing multiple operators that were not jointly optimized.
The experiments show that we can arrive at better accuracy by jointly fine-tuning these operators.
Furthermore, we find that our neural translation provides informative knowledge transfer from pre-trained pipelines, along with efficient network architecture that performs better than hand-designed networks with similar capacity.

\section{Pipeline Translation}
\label{sec:translation}

A machine learning pipeline is defined as a DAG of data-processing operators, and these operators are mainly divided into two categories: (1) the \textit{arithmetic} operators and (2) the \textit{algorithmic} operators.
Arithmetic operators are typically described by a single mathematical formula.
These operators are, in turn, divided into two sub-categories of \textit{parametric} and \textit{non-parametric} operators.
Non-parametric operators define a fixed arithmetic operation on their inputs; for example, the Sigmoid function can be seen as a non-parametric arithmetic operator.
In contrast, parametric operators involve numerical parameters on the top of their inputs in calculating the operators' outputs.
For example, an affine transform is a parametric arithmetic operator where the parameters consist of the affine weights and biases.
The parameters of these operators can be potentially tuned via some training procedure.
The algorithmic operators, on the other hand, are those whose operation is not described by a single mathematical formula but rather by an algorithm.
For instance, the operator that converts categorical features into one-hot vectors is an algorithmic operator that mainly implements the look-up operation.

Given a DAG of arithmetic and algorithmic operators, we propose the following general procedure for translating it into a single neural network:
\begin{enumerate}
    \item \label{step:arithmetic} For an arithmetic operator, translate the mathematical formula into a neural network module.
    In the case of parametric operator, copy the values of the operator's parameters into the resulting neural module.
    \item \label{step:algorithmic} For an algorithmic operator, translate the operator by rewriting the algorithm as a differentiable module (if possible), or keep it as is.
    \item Compose all the resulting modules from Step~\ref{step:arithmetic} and~\ref{step:algorithmic} into a single neural network by following the dependencies in the original pipeline.
\end{enumerate}

The final output of the above translation process is a neural network that provides the same prediction results (unless the translation includes approximation) as the original pipeline on the inputs.
Note that Step~\ref{step:arithmetic} and~\ref{step:algorithmic} in the above procedure are where the actual translation happen, which are described in details next.

\subsection{Translating Arithmetic Operators}
\label{sec:arithmetic}
It is straightforward to translate a non-parametric arithmetic operator into a neural network module: the mathematical function of the operator can in fact be directly rewritten using the math API provided by a neural network framework.
On the other hand, parametric arithmetic operators are often implicitly derived from ML models\footnote{Note that some parametric operators exist that are not derived from ML models (e.g., normalizers). These operators, however, can be translated with the same mechanism used for parametric operators of ML models.}, which are not straightforward to translate.
ML models typically consist of three key components: (1) the prediction function, (2) the loss function, and (3) the learning algorithm.
While the prediction function defines the functional form of the model, the learning algorithm and the loss function define how it is trained toward what objective, respectively.
Take the popular linear Support Vector Machine (SVM) model as an example: the prediction function is a linear function of certain input dimensionality; the loss function is the Hinge loss, and the learning algorithm is gradient descent in the dual space.

A crucial observation is that once the training is complete, the data-processing operation of any ML model can be completely defined by the prediction function regardless of the loss function and the learning algorithm.
Hence, we can translate these parametric operators by applying the translation method for non-parametric operators to their prediction functions.
For example, a linear SVM model can be translated into a linear layer of one output unit having the weights transferred from the parameters of the trained model.
It is worth noting that the translation of a trained ML pipeline into a neural network is uniquely done starting from the prediction function, independently on how different parts of it have been trained.
This is a powerful observation because it enables us to translate different operators of a pipeline using the same formalism even though they might have been obtained via different learning algorithms or objectives.

\subsection{Translating Algorithmic Operators: Decision Trees}
\label{sec:tree}

\begin{figure*}[t]
\centering
\begin{subfigure}{0.34\textwidth}
    \centering
	\includegraphics[width=1.0\textwidth]{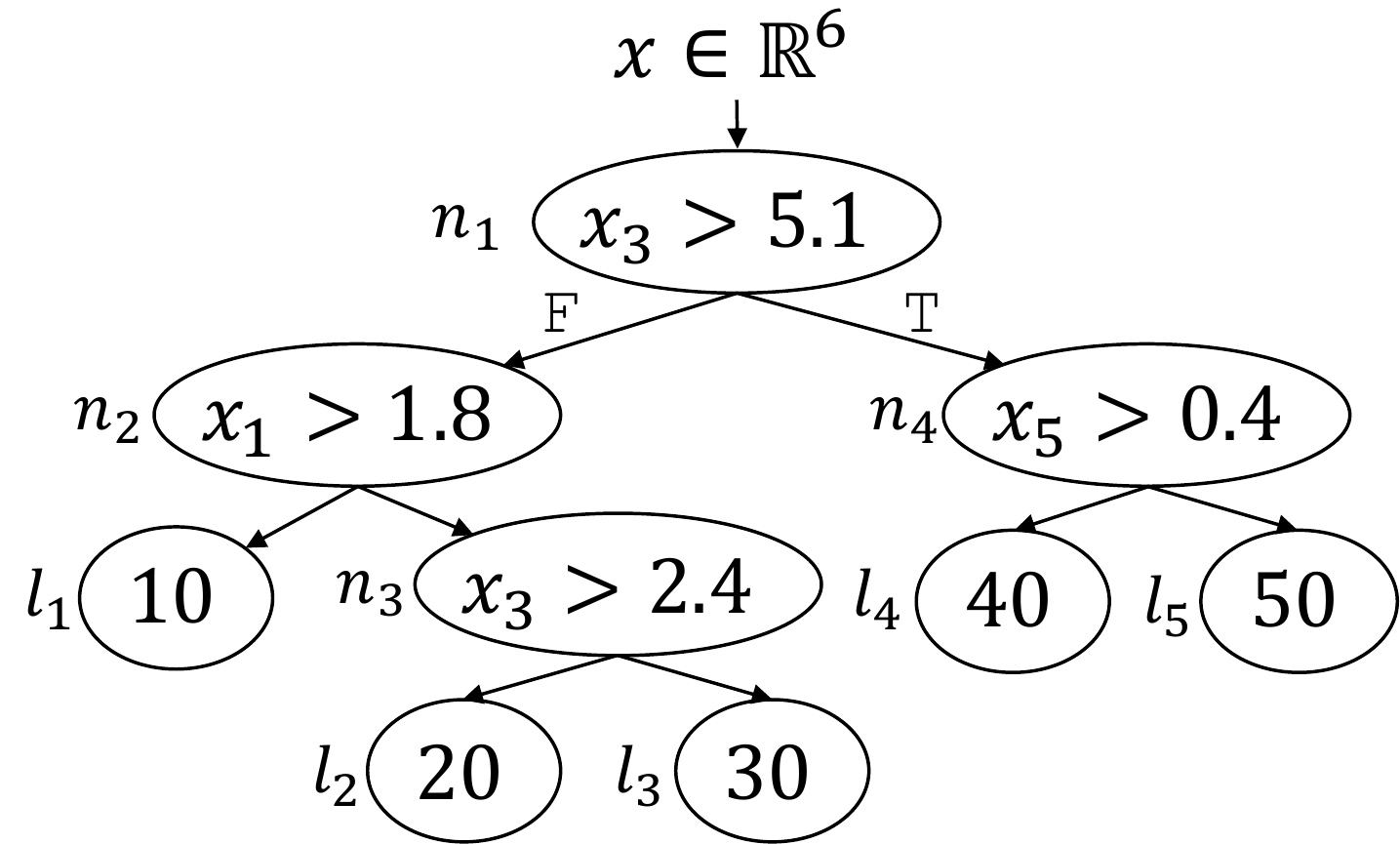}
	\centering
	\caption{An example decision tree.}
	\label{fig:decision-tree}
\end{subfigure}
\hspace{0.01\textwidth}
\begin{subfigure}{0.25\textwidth}
    \centering
	\includegraphics[width=1.0\textwidth]{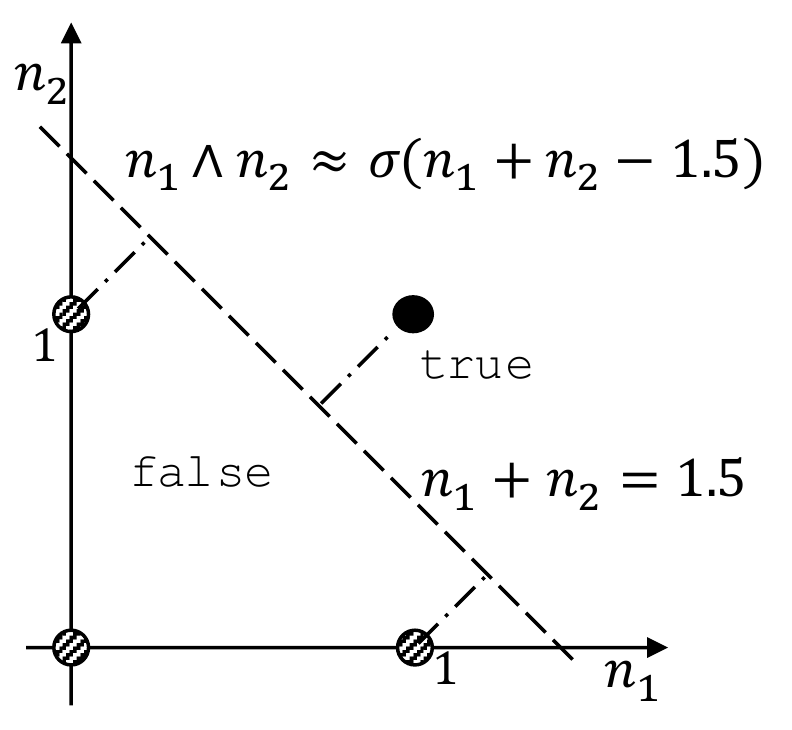}
	\centering
	\caption{Expressing logical conjunction using arithmetic operations.}
	\label{fig:conjunction}
\end{subfigure}
\hspace{0.01\textwidth}
\begin{subfigure}{0.29\textwidth}
    \centering
	\includegraphics[width=1.0\textwidth]{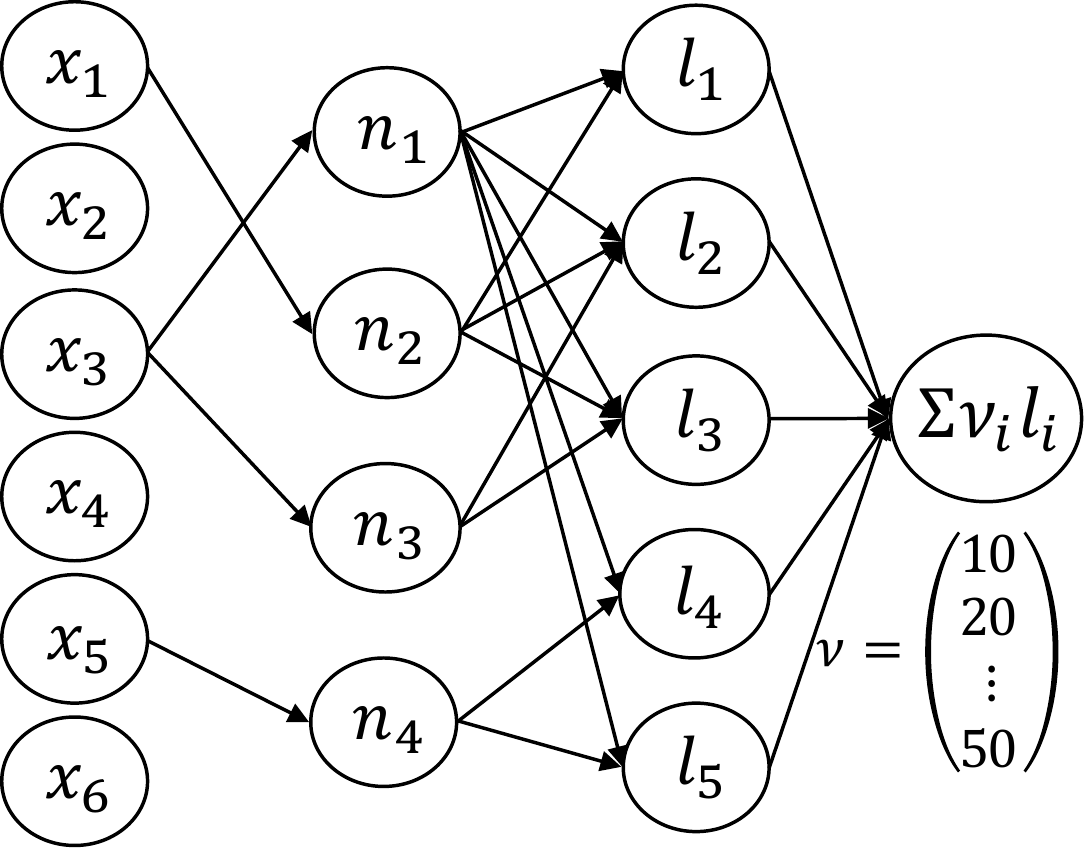}
	\centering
	\caption{A neural network translated from the decision tree of Figure~\ref{fig:decision-tree}.}
	\label{fig:translating-tree}
\end{subfigure}
\caption{Translating a decision tree into a multi-layer perceptron.}
\end{figure*}

While most ML models produce differentiable arithmetic operators that can be directly translated, some do not.
Among such models are the popular tree models whose prediction functions (i.e. the decision trees) are not a simple differentiable function.
Instead, each prediction is made by executing a sequence of \texttt{if-else} statements.
In that respect, a tree prediction function is an algorithmic operator rather that an arithmetic one.
During translation, we can treat it as such and simply translate a decision tree as a nested \texttt{if-else} statements, however, the main problem with this approach is that we will not be able to \textit{parametrize} the tree prediction function and further fine-tune it.
In order to do so, we would need to rewrite the tree prediction function as an arithmetic operator instead, which is not trivial.

To tackle this challenge, we take inspiration from an early work~\cite{Banerjee94initializingneural} that initializes a Multi-Layer Perceptron (MLP) from a decision tree.
At a given internal node $n$ of a binary decision tree, the tree prediction algorithm evaluates the decision function $d_n(x)=\mathbb{I}(x_{i(n)} > \theta_n)$, where $x$ is a vector representing the input of the tree, $i(n)$ is the index of the feature examined at node $n$, $\theta_n$ is the decision threshold at node $n$, and $\mathbb{I}(\cdot)$ is the indicator function.
We replace this non-differentiable function with a smooth one $d_n(x) \approx \sigma(x^Te_{i(n)} - \theta_n)$ following Banerjee's,
where $e_{i(n)}$ is the canonical basis vector along the $i(n)$-th dimension of the feature space and $\sigma(\cdot)$ is the Sigmoid function.
The difference from Banerjee's is that we create one neuron for each internal node while they create two.
Therefore they lead to more redundant network that requires twice more computation to represent the same tree prediction function.

Next, we note that the value of a leaf node is outputted as the final value iff the path from the root node to that leaf node is traversed.
For example, in Figure~\ref{fig:decision-tree}, the tree will output $40$ (i.e. the value of leaf $l_4$) iff $d_{n_1}(x)=1$ and $d_{n_4}(x)=0$.
As such, we denote the leaf activation function of $l_4$ as a conjunction of decision functions of $l_4$'s ancestors: $l_4(x) = d_{n_1}(x) \wedge \neg d_{n_4}(x)$.
To get a differentiable approximation of the logical conjunction, we can write $\bigwedge_{i=1}^C a_i \approx \sigma\bigg(\sum_{i=1}^C \mu(a_i) - C + 0.5\bigg)$ where $C$ is the total number of literals in the conjunction (the path length from the root to the leaf) and $\mu(a)=\begin{cases} 1-d  & \mbox{if } a = \neg d \\ d & \mbox{otherwise} \end{cases}$.
Figure \ref{fig:conjunction} visualizes this approximation for 2 inputs.
The equation $n_1 + n_2 = 1.5$ is a maximum-margin hyperplane between \texttt{true} and \texttt{false} evaluations.

Having translated the basic operations of a tree prediction function into smooth functions as above, any decision tree can be translated into a MLP with two hidden layers.
Figure~\ref{fig:translating-tree} shows an example of this translation procedure.
The first hidden layer implements a hidden unit ($d_n(\cdot)$) per each internal node.
The second hidden layer allocates a hidden unit ($l_n(\cdot)$) for each leaf node.
Finally, the output layer is defined as a linear layer with one unit, $T(x)=\sum_{i\in L}\nu_i l_i(x)$,
where $L$ is the set of all leaf nodes and $\nu_i$ is the value of the leaf node.
Note that, in the case of no approximation, one and only one of the leaf activation functions $l_i(x)$ evaluates to $1$ for any given input $x$, while the rest are $0$.
The just described technique can be generalized over tree ensembles, and in fact in the experimental section we will evaluate it over a grandient boosting model.

Once the translation of a tree is complete, the main question is which of the parameters of the resulting neural network should be declared as trainable.
We suggest four levels of parametrization to balance good fit and inductive bias:
\begin{enumerate}
    \item The leaf node values $\nu$ that constitute the weights of the output layer are declared as trainable.
    \item In addition to $\nu$'s, the decision threshold values $\theta$'s at the internal nodes are declared as trainable. These parameters constitute the bias parameters of the first hidden layer.
    \item In addition to Level 2's parameters, the canonical basis vectors $e_{i(n)}$ in the equation of $d_n(\cdot)$ are replaced by a vector of free parameters of the same size.
    These parameters constitute the weights of the first hidden layer.
    \item In addition to Level 3's parameters, all the weights (including the non-existing $0$ weights) of the second hidden layers are declared as trainable.
\end{enumerate}

As level number increases, we declare more parameters as trainable and as such increase the capacity of the resulting neural network to fit to data better.
While Levels $1$ and $2$ can only change the leaf and the decision threshold values in the tree, Level $3$ can additionally lead to examining a linear combination of features at each internal node rather than a single feature.
Up to Level $4$, the tree structure is preserved; whereas, at Level $4$, we let the entire decision structure of the tree change.
That is, Level $4$ gives us a fully-connected and fully-trainable MLP initialized by a (trained) tree.

\subsection{Translating Algorithmic Operators for Categorical Features}

\begin{figure}[t]
\centering
\includegraphics[width=0.6\textwidth]{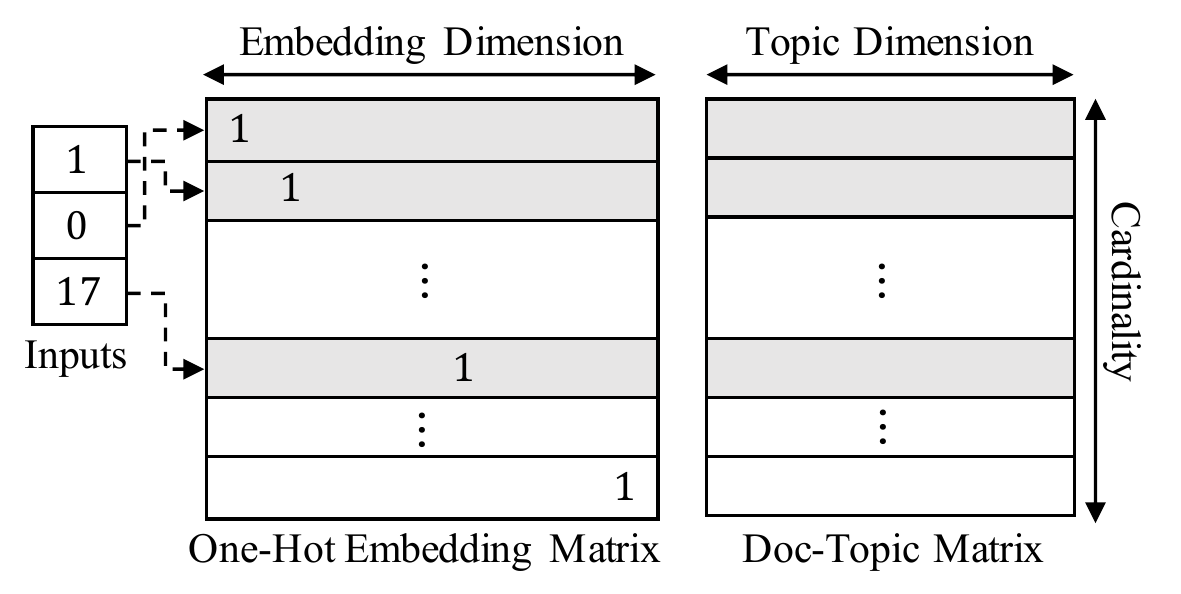}
\caption{Translating one-hot encoding and Latent Dirichlet Allocation into embedding lookup modules.}
\label{fig:onehot}
\end{figure}

Classical ML pipelines often convert categorical features into numerical values using non-differentiable, algorithmic operators.
The simplest yet most popular technique is one-hot (hash) encoding, which generates sparse one-hot vectors out of categorical inputs.
This operator is intrinsically non-differentiable since its inputs lie on discrete spaces such as integer or string.

Our observation is that one-hot encoding consumes raw categorical inputs, which means that we do not have to backpropagate further through its discrete inputs due to the absence of upstream operators.
We mainly rely on the embedding technique that has been studied intensively by deep learning community.
As shown in Figure~\ref{fig:onehot}, one-hot encoding can be seen as an embedding vector lookup operation with the embedding dimension matching the cardinality of categories.
We can declare this embedding matrix as trainable in order to replace sparse one-hot vectors with dense representations and learn relationships between different features, which is not possible with one-hot encoding.
The same statement holds for hash encoding, except that we can make the embedding dimension much smaller than the cardinality because it uses the hashing trick.

In addition to simply converting categorical features into numerics, we may want to leverage interactions between features.
For example, given two categorical features such as user id and product id, one will try to extract information from interaction between these features using decomposition techniques such as matrix factorization or Latent Dirichlet Allocation (LDA)~\cite{lda}.
LDA, in particular, is a generative probabilistic model that learns abstract topics from a collection of documents.
LDA's prediction function typically performs variational inference or Gibbs sampling to produce a document-topic assignment vector.
Hence, the LDA prediction function is a non-differentiable algorithmic operator.

When we incorporate LDA within ML pipelines, we use the document-topic vectors produced by LDA module as features for downstream operators.
Such prediction function is again similar to the embedding lookup: for each categorical feature (document), we calculate a dense representation of the feature (document-topic distribution).
This behavior allows us to translate LDA into an embedding-lookup operation with a pre-calculated document-topic matrix as described in Figure~\ref{fig:onehot}.
We can denote the document-topic matrix as trainable parameters to reach a better optimum by employing the final loss for updating the matrix, which was originally obtained by unsupervised procedure without explicit signal from the prediction target.

\subsection{Limitations}
Unfortunately, there are some algorithmic operators that we cannot translate into a differentiable format yet.
Word tokenization and missing data imputation are such examples.
Since our translation approach currently do not handle these operators, we do not translate them and keep them as they are.
Nevertheless, in all the cases we studied, these non-translatable operators are placed at the beginning of the pipeline and do not affect backpropagation through the rest of the translated network.
Hence, we can still compute gradients and fine-tune the downstream operators, which are the more essential parts of the original ML pipeline.

\subsection{Fine-Tuning}
\label{sec:fine-tuning}

After translating the ML pipeline into a neural network, one can further fine-tune the trainable parameters of the resulting network via backpropagation.
There are many scenarios for which this fine-tuning step can be useful.
First, by fine-tuning the resulting network on the original training data, we can potentially improve the generalization of the model since we are now jointly optimizing all the operators of the pipeline toward the final loss function.
Second, as we discussed above, the translation process does not depend on the loss functions different operators of the pipeline have been trained toward before.
This means that once the translation is complete, the resulting network can be fine-tuned toward a completely different objective that is more suitable for a given application.
Third, fine-tuning can be used to adapt the model to new data that were not available before, which is not straightforward without re-training the original ML pipeline with the old and new data.
It is worth noting that other methods for fine-tuning such as boosting may increase the model size and complexity, while our translation approach does not.
Also, the ensemble model obtained by boosting can be seen as a pipeline containing multiple operators that were not jointly optimized, so it can also benefit from our translation approach.

\section{Experiments}
\label{sec:eval}

In this section, we empirically evaluate the performance of our translation approach.
The main goal of the experiments is to show the followings: (1) we can improve the performance of ML pipelines by employing backpropagation instead of training each operator individually; (2) the translation of trained ML pipelines provides informative initialization of neural networks; and (3) the translation provides efficient neural architectures.
We carry our experiments on a binary classification task by using three tabular datasets under two different scenarios to showcase the capabilities our translator is able to provide.

\begin{table}[t]
\centering
\caption{
Model sizes for all network configurations and datasets.
We use S1-L\# to indicate the network translated from scenario 1 with optimization level \#.
S1-OHE and S1-LDA are the configurations in which we enable fine-tuning of one hot encoding and LDA, respectively.
MLP-1 and MLP-2 indicate the MLP baselines that roughly matches the number of parameters of S1-L4 and S2 (scenario 2), respectively.
Note that S1-OHE of Criteo uses less parameters compared to S1-L4 because we use less number of bits for feature hashing to prevent the embedding dimension from getting too large.
}
\label{tbl:model-size}
\begin{tabular}{cccc}
\toprule
Networks & Credit & Flight & Criteo \\ \midrule
S1-L1 & 2.5E+3 & 3.0E+3 & 3.0E+3 \\
S1-L2 & 4.9E+3 & 5.9E+3 & 5.9E+3 \\
S1-L3 & 3.4E+6 & 2.0E+6 & 4.8E+7 \\
S1-L4 & 3.5E+6 & 2.1E+6 & 4.8E+7 \\
S1-OHE & N/A & 2.3E+6 & 5.0E+6 \\
S1-LDA & N/A & 7.6E+5 & N/A \\
MLP-1 & 2.5E+6 & 1.8E+6 & 3.8E+7 \\
DeepGBM & N/A & 7.4E+5 & 2.1E+7 \\
S2 & N/A & 3.8E+4 & 5.3E+4 \\
MLP-2 & N/A & 4.9E+4 & 6.6E+4 \\ \bottomrule
\end{tabular}
\end{table}

\paragraph{Datasets.}
The Credit\footnote{\url{https://www.kaggle.com/c/home-credit-default-risk/data}} dataset includes around 310K records and aims at predicting whether clients are capable of paying back their loans.
Each record contains 209 features in total: among them 172 are numeric, and 37 are categorical.
Training, validation and test datasets are carved from the full dataset after a shuffling step and contain 250K, 30K, and 30K records respectively.

The Flight\footnote{\url{http://stat-computing.org/dataexpo/2009/}} dataset includes around 21M records and is used for predicting whether a scheduled flight will be delayed (more than 15 minutes).
Each record has 8 features, where 2 are numeric and 6 are categorical.
For the set of experiments using Flight, we use years 2006 and 2007 as training set (about 14M records), while year 2008 is divided in 2 and used as validation and test set (about 3.4M records each).

The Criteo\footnote{\url{https://labs.criteo.com/2014/02/kaggle-display-advertising-challenge-dataset/}} dataset aims at predicting the click-through rate for an online advertisement.
It includes around 46M records each of them with 39 features.
Among the 39 features, 13 are numeric while the remaining are categorical.
Criteo requires hashing trick due to the high cardinality of categorical features.
Training, validation and test datasets, each of them containing respectively, 44M, 1M, and 1M records, are carved from the full dataset after a shuffling step.

\paragraph{Scenarios.}
We evaluate the performance improvements unlocked by our neural translation approach through two scenarios.
In the first scenario, we use a pipeline employing a tree ensemble trained by LightGBM~\cite{lgbm}.
Categorical features are either handled by (1) one-hot encoding (with hashing in case of Criteo) or (2) Latent Dirichlet Allocation (only experimented on Flight).
We measure the performance of the neural translation with different tree parametrization levels (L1 to L4).
Furthermore, we experiment with two regimes of network initialization for the parameters that are declared as trainable:
(1) in the \textit{warm start} regime the parameter values are carried over from the trained ML pipeline (denoted by ``Warm'' in the result tables); and (2) in the \textit{cold start} regime the trainable parameters are randomly initialized (denoted by ``Cold''), while the other parameters not declared as trainable are transferred intact from the original pipeline.
Since different levels of parametrization for tree translation introduce different trade-offs between good fit and strong inductive bias, we report results on different training sample sizes.

For the second scenario, we compose the pipeline as follows: (1) apply Principal Component Analysis (PCA) to the input feature $x$ and produce $\tilde{x}$, which resides in the principal component space of $x$; (2) train a LightGBM model using $\tilde{x}$ and the input label $y$; (3) using the leaf activation function for each tree in the trained LightGBM model, create a one-hot vector that marks the index of the activated leaf as 1 and keeps others 0; (4) concatenate the output from (3) and the original input $x$; (5) train the final linear classifier model using the concatenated feature from (4) and the label $y$.
We train the final linear model using Stochastic Dual Coordinate Ascent~\cite{sdca}.
Due to the dimensionality reduction done by PCA, this scenario employs fewer parameters compared to the first scenario.
Categorical features are converted into one-hot vectors (with hashing in case of Criteo) before being fed to the PCA or the linear classifier.
Again, we experiment with both warm and cold start settings.

\paragraph{Baselines.}
In our experiments we compare against three baselines: (1) the original ML pipeline (denoted by ``ML'' in the tables), (2) a Multi-Layer Perceptron with 2 hidden layers (denoted by ``MLP''), and (3) DeepGBM~\cite{ke2019deepgbm}, a recent work that distills neural network from decision trees.
The MLP uses ReLU as an activation function, and employs dropout ratio of 0.1 on each hidden layer.
The MLP solution is designed such that it approximately matches the number of trainable parameters found in the network generated by the neural translation.
Table~\ref{tbl:model-size} shows the number of parameters for the translated neural networks and baselines (MLP and DeepGBM) used throughout the experiments.
More details about the experiment settings such as number of trees and batch size can be found in Appendix~\ref{apdx:exp_settings}.

\begin{table*}[t]
\centering
\caption{AUC of scenario 1 obtained by the ML pipeline, MLP, and translated network at different parametrization levels.}
\label{tbl:scenario-1-opt-level}
{
\notsotiny
\tablecolumnmarginsmall
\begin{tabular}{ccllllllllll}
\toprule
\multirow{2}[3]{*}{Dataset}  & \multirow{2}[3]{*}{ML}   & \multirow{2}[3]{*}{MLP}   & \multicolumn{2}{c}{Level 1}   & \multicolumn{2}{c}{Level 2}   & \multicolumn{2}{c}{Level 3}   & \multicolumn{2}{c}{Level 4}   \\
\cmidrule(lr){4-5} \cmidrule(lr){6-7} \cmidrule(lr){8-9} \cmidrule(lr){10-11}
& & & \multicolumn{1}{c}{Warm} & \multicolumn{1}{c}{Cold} & \multicolumn{1}{c}{Warm} & \multicolumn{1}{c}{Cold} & \multicolumn{1}{c}{Warm} & \multicolumn{1}{c}{Cold} & \multicolumn{1}{c}{Warm} & \multicolumn{1}{c}{Cold}  \\ \midrule
Credit & 0.7921 & 0.5960 & \textbf{0.7951} & 0.7866 & 0.7732 & 0.7567 & 0.6185 & 0.6204 & 0.6182 & 0.6215 \\
Flight & 0.7447 & 0.7196 & 0.7468 & 0.7464 & 0.7244 & 0.6122 & \textbf{0.7815} & 0.7728 & 0.7698 & 0.7525 \\
Criteo & 0.7756 & 0.7971 & 0.7833 & 0.7833 & 0.7571 & 0.7514 & \textbf{0.8036} & 0.8031 & \textbf{0.8036} & 0.8029 \\ \bottomrule
\end{tabular}
}
\end{table*}

\subsection{Scenario 1}

In scenario 1, we evaluate different tree parametrization levels (L1 to L4) using Credit, Flight, and Criteo dataset.
We use one-hot (hash) encoding for handling categorical features, if not otherwise stated.
The overall results can be found in Table~\ref{tbl:scenario-1-opt-level}.

From these results, we can see that:
\begin{itemize}
    \item The warm start outperforms the cold start (except a few cases with L3 and L4 on Credit), which means that the weights transferred from a pre-trained ML pipeline provide an informative initialization for the neural network.
    \item Joint optimization (by neural translation) of the decision trees within the ensemble improves the AUC over the original pipeline.
    \item As the training dataset gets bigger (Credit $<$ Flight $<$ Criteo), higher parametrization levels give the best results.
    This trend clearly shows slight overfitting of L3 and L4 in small data experiments and how it is avoided by L1 that has much fewer trainable parameters.
    In other words, lower levels provide a natural regularization mechanism in small data experiments.
    \item For L3 and L4 (except the Credit dataset that is too small) the cold start regime outperforms the ML or MLP baseline.
    This shows that the neural translation not only deliver meaningful information by transferring weights, but also provides a good network architecture that can achieve better results than the baselines.
\end{itemize}

\begin{table*}[t]
\centering
\caption{AUC of scenario 1 obtained by sampling the training dataset using different sizes.}
\label{tbl:scenario-1-data-size}
{
\notsotiny
\tablecolumnmarginsmall
\begin{tabular}{cccllllllll}
\toprule
\multirow{2}[3]{*}{Dataset} & \multirow{2}[3]{*}{Sample Size}  & \multirow{2}[3]{*}{ML}   & \multicolumn{2}{c}{Level 1}   & \multicolumn{2}{c}{Level 2}   & \multicolumn{2}{c}{Level 3}   & \multicolumn{2}{c}{Level 4} \\
\cmidrule(lr){4-5} \cmidrule(lr){6-7} \cmidrule(lr){8-9} \cmidrule(lr){10-11}
& & & \multicolumn{1}{c}{Warm} & \multicolumn{1}{c}{Cold} & \multicolumn{1}{c}{Warm} & \multicolumn{1}{c}{Cold} & \multicolumn{1}{c}{Warm} & \multicolumn{1}{c}{Cold} & \multicolumn{1}{c}{Warm} & \multicolumn{1}{c}{Cold} \\ \midrule
\multirow{5}{*}{Flight} & 1\% & \textbf{0.7335} & \textbf{0.7335} & 0.7276 & 0.7196 & 0.6044 & 0.7295 & 0.7195 & 0.7315 & 0.7198 \\
& 10\% & 0.7429 & 0.7441 & 0.7415 & 0.7277 & 0.6141 & 0.7525 & 0.7302 & \textbf{0.7544} & 0.7285 \\
& 30\% & 0.7421 & 0.7420 & 0.7412 & 0.7298 & 0.6148 & 0.7616 & 0.7383 & \textbf{0.7676} & 0.7487 \\
& 50\% & 0.7425 & 0.7426 & 0.7416 & 0.7292 & 0.6151 & 0.7695 & 0.7579 & \textbf{0.7778} & 0.7686 \\
& 70\% & 0.7420 & 0.7412 & 0.7390 & 0.7298 & 0.6158 & 0.7774 & 0.7667 & \textbf{0.7806} & 0.7762 \\ \midrule
\multirow{5}{*}{Criteo} & 1\% & 0.7704 & \textbf{0.7717} & 0.7687 & {0.7196} & {0.7007} & {0.7678} & {0.7698} & {0.7680} &{0.7697} \\
& 10\% & 0.7748 & 0.7818 & 0.7816 & {0.7410} & {0.7285} & \textbf{0.7852} & 0.7845 & 0.7849 & 0.7847 \\
& 30\% & 0.7756 & 0.7832 & 0.7831 & {0.7485} & {0.7401} & \textbf{0.7950} & 0.7913 & 0.7947 & 0.7920 \\
& 50\% & 0.7752 & 0.7830 & 0.7830 & {0.7526} & {0.7455} & \textbf{0.7991} & 0.7973 & \textbf{0.7991} & 0.7980 \\
& 70\% & 0.7753 & 0.7831 & 0.7831 & {0.7539} & {0.7476} & 0.8016 & 0.8003 & \textbf{0.8018} & 0.8006 \\ \bottomrule
\end{tabular}
}
\end{table*}

\begin{table}[t]
\centering
\caption{
AUC of scenario 1 obtained by fine-tuning the translated network (Level $4$) with dropout (L4 + Dropout) and further declaring one-hot encoding as trainable (+ OHE).}
\label{tbl:scenario-1-final}
\begin{tabular}{ccccccc}
\toprule
 & \multirow{2}[3]{*}{ML} & \multirow{2}[3]{*}{DeepGBM} & \multicolumn{2}{c}{L4 + Dropout} & \multicolumn{2}{c}{+ OHE} \\
\cmidrule(lr){4-5} \cmidrule(lr){6-7}
& & & \multicolumn{1}{c}{Warm} & \multicolumn{1}{c}{Cold} & \multicolumn{1}{c}{Warm} & \multicolumn{1}{c}{Cold} \\ \midrule
Flight & 0.7447 & 0.7996 & 0.7875 & 0.7629 & \textbf{0.8067} & 0.7927 \\
Criteo & 0.7756 & 0.8092 & 0.8045 & 0.8023 & \textbf{0.8102} & 0.8052 \\ \bottomrule
\end{tabular}
\end{table}

\begin{table}[t]
\centering
\caption{
AUC of scenario 1 for Flight where we replaces some one-hot encoding modules with LDA.
The results are obtained by fine-tuning the translated network with dropout and with LDA set as trainable.
}
\label{tbl:scenario-1-lda}
\begin{tabular}{cccc}
\toprule
 & \multirow{2}[3]{*}{ML} & \multicolumn{2}{c}{L4 + Dropout + LDA} \\
\cmidrule(lr){3-4}
 & & \multicolumn{1}{c}{Warm} & \multicolumn{1}{c}{Cold} \\ \midrule
Flight & 0.7462 & \textbf{0.8051} & 0.7861 \\ \bottomrule
\end{tabular}
\end{table}

We additionally report results on different training sample sizes in Table~\ref{tbl:scenario-1-data-size}.
For this experiment we skip the Credit dataset, since it is already small (only Level $1$ is able to improve performance compared to the original ML pipeline).
The results further support the claim that lower parametrization levels provide natural regularization.
In addition, we try adding explicit regularization to Level $4$.
Dropout with a zeroing probability of $0.1$ is applied at the second hidden layer of the network described in Fig.~\ref{fig:translating-tree}.
The results can be found in Table~\ref{tbl:scenario-1-final}.
By comparing with the results from Table~\ref{tbl:scenario-1-opt-level}, we can observe that in this case, Level $4$ with dropout is the best strategy for training the tree part of the network.

We further declare the modules for handling categorical features as trainable on top of Level $4$ with dropout, and the results can be found in the last column of Table~\ref{tbl:scenario-1-final}.
By further jointly optimizing the one-hot encoding module with the networks generated from the trees, we can further improve the performance.
In particular, by compressing the redundant sparse features into dense embeddings, the Criteo pipeline (S1-OHE in Table~\ref{tbl:model-size}) uses almost 10 times less parameters compared to the pipeline that does not declare one-hot encoding as trainable (S1-L4 in Table~\ref{tbl:model-size}), but reaches better performance.
Our solution also produces better results compared to DeepGBM~\cite{ke2019deepgbm}, a recent tree translation work, that employs comparable amount of parameters.
We attribute the gap between DeepGBM and our solution to the joint optimization of multiple components (in this case, one-hot encoding and LightGBM) via backpropagation.

Finally, in addition to the scenario 1 pipelines that uses one-hot encoding, we experiment with another scenario 1 pipeline that uses LDA instead of some one-hot encoding modules for handling the categorical features of the Flight dataset.
Table~\ref{tbl:scenario-1-lda} reports the numbers for this experiment.
While the ML pipeline with LDA performs slightly better than the pipeline with one-hot encoding due to the ineffectiveness of LightGBM in handling sparse features, the results of translated network are reversed.
This implies that the network from pipeline with one-hot encoding somehow leverages the opportunity for more flexible fit provided by its larger model capacity (see S1-OHE and S1-LDA of Table~\ref{tbl:model-size}).

\begin{table}[t]
\centering
\caption{AUC of scenario 2 obtained by the original ML pipeline (ML), MLP, and translated network.}
\label{tbl:scenario-2}
\begin{tabular}{ccccc}
\toprule
\multirow{2}[3]{*}{Dataset} & \multirow{2}[3]{*}{ML} & \multirow{2}[3]{*}{MLP} & \multicolumn{2}{c}{L4 + Dropout} \\
\cmidrule(lr){4-5}
& & & \multicolumn{1}{c}{Warm} & \multicolumn{1}{c}{Cold} \\ \midrule
Flight & 0.6990 & 0.7223 & \textbf{0.7284} & 0.7082 \\
Criteo & 0.7644 & 0.7793 & \textbf{0.7904} & 0.7903 \\ \bottomrule
\end{tabular}
\end{table}

\subsection{Scenario 2}

For the experiments in this scenario we use only the Flight and Criteo datasets.
The purpose of this experiment is to test our neural translation approach on classical ML pipelines composed of several distinct models.
We use Level $4$ with dropout for fine-tuning the tree part of the translated network, which was the best strategy in scenario 1.
Table~\ref{tbl:scenario-2} reports the result of fine-tuning, compared to the ML and MLP baselines.
Similarly to the previous set of experiments, we can observe that fine tuning the translated network improves the AUC compared to the baselines.
One interesting result to notice is that in case of Flight dataset, the MLP baseline with less parameters (MLP-2) performs better than the MLP with more parameters (MLP-1).
This shows that neural networks with larger capacity do not always lead to better results, while our translation approach can bypass this hurdle by using informative initialization (Table~\ref{tbl:scenario-1-opt-level}, L3 and L4 on Flight dataset).

\section{Related Works}
\label{sec:rltwk}

Milutinovic et al.~\cite{e2e} proposes the end-to-end training of ML pipelines via propagating gradients across multiple differentiable operators (possibly from different libraries).
This work however has no discussion about non-differentiable operators, while we attempt to backpropagate through non-differentiable (decision tree and LDA) and non-trainable operators (one-hot encoding) and transfer knowledge from the pre-training phase.
Additionally, this work requires users to manually write ``backward'' code for operators from non-NN libraries (e.g., scikit-learn), while our neural translation approach exploits the automatic differentiation capabilities of neural network libraries.

As we mentioned earlier, there has been early work~\cite{Banerjee94initializingneural} that tries to translate a pre-trained decision tree into a MLP, and we took inspiration from this.
Ivanova and Kubat~\cite{Ivanova95initializationof} suggested similar approach, however, they carry out the computation that corresponds to our first hidden layer in Figure~\ref{fig:translating-tree} outside the network and cannot backpropagate further.
Humbird, Peterson, and McClarren~\cite{8478232} uses a decision tree to initialize a MLP in a different way, where the depth of the decision tree is used to decide the number of layers.
Weights are randomly initialized, while the information on the tree is retained only for sparsely connecting the neurons.
We instead use all the information of pre-trained trees including but not limited to the tree structure and decision thresholds to initialize the network weights.
Yang,  Morillo,  and  Hospedales~\cite{dndt} suggests to build tree-like neural networks for interpretability.
This is different from our approach of handling trees because: (1) it builds a tree-like neural network using random weight initialization, while we retain the behavior of pre-trained trees by neural translation; and
(2) our translated network learns to use all features for making decision at each internal node (L3 and L4), while this work uses a single feature at each neuron and requires a wider network whose number of neurons grows exponentially as the number of features grows.
Kontschieder  et  al.~\cite{dndf} combines neural network and decision trees by enabling backpropagation.
However, this work mainly focused on computer vision tasks without attention to tabular data that require other algorithmic operators to handle categorical features.
It also maintains the tree structure (i.e., the path from the root to leaf) while we may let the structure change and exploit all internal nodes (L4).
Finally, DeepGBM~\cite{ke2019deepgbm} \textit{distills} decision trees into neural networks by transferring the knowledge of tree outputs and feature importance learned by gradient boosting trees.
Given this distilled neural network, DeepGBM incorporates an additional NN component called \textit{CatNN} for better handling of categorical input features.
We instead directly \textit{translate} the classical ML pipeline into neural network, which makes the translation process much more intuitive.
Also, our translation approach provides various neural network architectures by changing the structure of the original ML pipeline, while DeepGBM produces a fixed architecture.
This difference can make the translation approach more generally applicable.

Stoyanov, Ropson, and Eisne~\cite{erma} deal with algorithmic operator beyond trees.
Specifically, they directly minimizes the empirical risk (aka expected loss) of a Markov Random Field by backpropagating through the inference algorithm.
While this work deals with a system composed of a single model (Markov random field), our work handles pipelines with multiple operators.

Since the one-at-a-time training strategy of classical ML can be seen as a greedy approach, we can compare our approach with other works that alleviates wrong decisions due to the greedy procedure.
Goyal, Dyer, and Berg-Kirkpatrick~\cite{diffss} suggests to backpropagate through the greedy scheduled sampling technique~\cite{schedule} in order to help to correctly assigning credits for errors.
In contrast, our work not only mitigates the greedy behavior of ML pipelines by backpropagation but also transfers knowledge from pre-trained pipelines into neural networks.
Our experiments show that these two contributions combined together (``warm start") and outperform the ``cold start'' training of the translated neural network (backpropagation alone without knowledge transfer).

\section{Conclusions}
\label{sec:conc}

Inspired by the existing gap between classical ML pipelines and neural networks, in this paper, we propose a framework for translating ML pipelines into neural networks and further \textit{jointly} fine-tuning them.
As part of our translation procedure, we also propose techniques for translating popular non-differentiable operators including decision trees and one-hot encoding.
The experimental results show that the translation with knowledge transfer followed by the fine-tuning leads to significant accuracy improvements over the original pipeline and hand-designed neural networks.
Furthermore, we see that our translation mechanism can be seen as an approach for designing neural network architectures for a given task that is inspired by the classical ML pipeline designed for that task.
We deem this work as a first step towards filling the gap between classical ML pipelines and neural networks.

\bibliographystyle{plain}
\small
\bibliography{finetune}

\begin{appendices}
\section{Experimental Settings}
\label{apdx:exp_settings}
Here we add more details about our experimental settings.

\begin{figure}[h]
\centering
\begin{subfigure}{0.27\textwidth}
    \centering
	\includegraphics[width=1.0\textwidth]{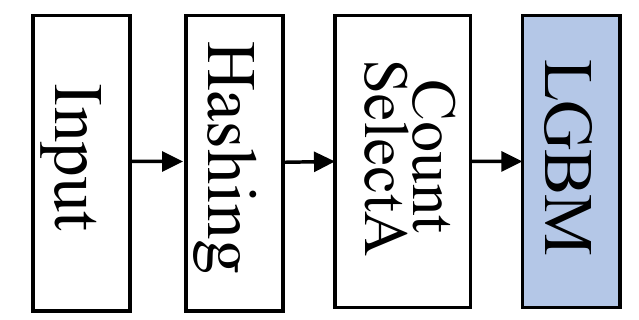}
	\centering
	\caption{Scenario 1.}
	\label{fig:scenario1}
\end{subfigure}
\hspace{0.07\textwidth}
\begin{subfigure}{0.481\textwidth}
    \centering
	\includegraphics[width=1.0\textwidth]{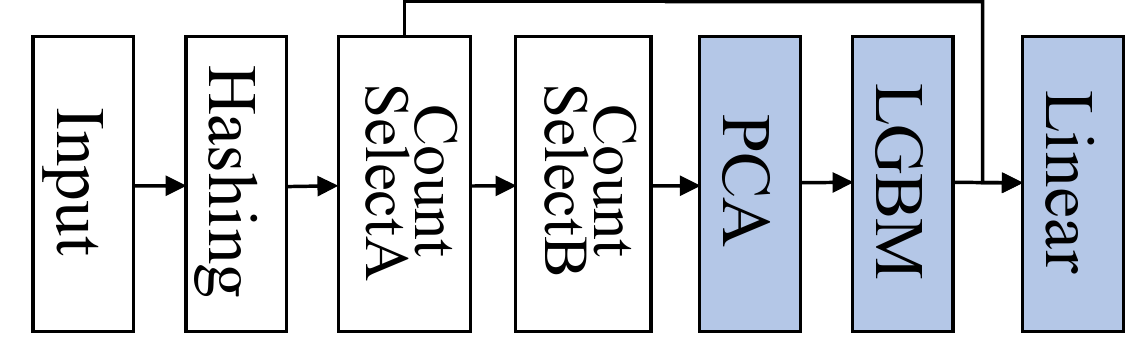}
	\centering
	\caption{Scenario 2.}
	\label{fig:scenario2}
\end{subfigure}
\caption{The scenarios used on the Criteo and Flight datasets.
The gray boxes represent the operators that are jointly optimized during the fine-tuning process.}
\label{fig:scenarios}
\end{figure}

\subsection{Criteo Dataset}

In this set of experiments we aim at predicting the click-through rate for an online advertisement.
We compose a ML pipeline that (1) fills in the missing values in the numerical columns of the dataset; (2) encodes categorical columns into one-hot vectors using a hash function with 10 bits (``Hashing'' in Fig.~\ref{fig:scenarios}); (3) discard feature dimensions that do not have any record with nonzero value (``CountSelectA'' in Fig.~\ref{fig:scenarios}); and (4) feeds the data into either scenario 1 (Fig.~\ref{fig:scenario1}) or scenario 2 (Fig.~\ref{fig:scenario2}).
For both scenarios, we set the LightGBM to create 30 leaves for each tree, while we constructed 100 and 30 trees in scenario 1 and 2, respectively.
In the experiment that declares one-hot encoding as trainable (S1-OHE in Table 1), we use 6 bits hashing to ensure that the embedding dimension of each categorical feature is no greater than 64.
The PCA transform used in scenario 2 follows ``CountSelectB'' that selects frequently occurring slots by using a threshold of 150K.
We use ML.NET default settings for the other hyperparameters.

Regarding the translated networks, we fine-tune them using the Adam optimizer with a batch size of 4096.
For scenario 1, we use a learning rate (lr) of 1e-5 and a weight decay (wd) of 1e-6, while for scenario 2, we use (lr, wd) = (1e-3, 1e-8) for the learning rate and the weight decay, respectively.
We select these hyperparameters by sweeping the space [1e-2, 1e-6] for lr and [1e-5, 1e-9] for wd, and using a fixed batch size of 4096.
We let the training process run until convergence.

\subsection{Flight Dataset}

In this second set of experiments our aim is to predict whether a scheduled flight will be delayed (more than 15 min) or not according to historical records.
We first convert all the categorical columns by using one-hot encoding (instead of the one-hot hash encoding ``Hashing'', and omit the count selectors ``CountSelectA'' and ``CountSelectB'' in Fig.~\ref{fig:scenarios}); successively, as for Criteo, we apply either scenario 1 or scenario 2.
When replacing one-hot encoding with LDA, we replace 3 one-hot encoding modules out of total 6 one-hot encoding modules (total 6 categorical features).
The LDA modules learn interaction between the selected 3 categorical features, while the other 3 one-hot encoding modules remain as is and declared as trainable when LDA modules become trainable.
The hyperparameters are the same as Criteo's, except that we use (lr, wd) = (1e-4, 1e-8) and (1e-4, 1e-6) for scenario 1 and 2, respectively, obtained from the parameter sweep.

\subsection{Credit Dataset}

In the last set of experiments, our aim is to predict whether clients are capable of paying back their loans. 
We first conduct feature engineering by introducing new hand-crafted features in order to improve the pipeline accuracy. 
By subtracting or dividing two different existing features, we are able to produce new meaningful features (i.e. credit-to-income-ratio) that improve the overall accuracy. 
Also if there are multiple values for a single feature, summaries of these values  (i.e. mean, variation, and summation) are introduced as new features. 
After that, we fill all the missing values with zeros and convert all the categorical features using one-hot encoding.
The final data is fed into scenario 1, where the LightGBM constructs 100 trees with 25 leaves each.

In case of the translated network, we use Adam optimizer with a batch size 256.
Hyperparameters are set as (lr, wd) = (1e-6, 1e-3) after sweeping the search space.

\subsection{MLP Baseline}
We use the Adam optimizer and a batch size equivalent to the one used for training the network output of our translator (4096 for Criteo and Flight; 256 for Credit). We select learning rate and weight decay by sweeping the same space as the translated network.

\subsection{DeepGBM Baseline}
We use an open-source implementation\footnote{\url{https://github.com/motefly/DeepGBM}} written by the authors of DeepGBM~\cite{ke2019deepgbm}.
The only difference from the experiments of the original DeepGBM paper is that we use our train-valid-test dataset split (described in Section~\ref{sec:eval}) for fair comparison.

\end{appendices}

\end{document}